\documentclass[10pt,twocolumn,letterpaper]{article}

\usepackage{cvpr}              

\usepackage[dvipsnames]{xcolor}

\usepackage{multirow}
\usepackage{mathrsfs}

\definecolor{cvprblue}{rgb}{0.21,0.49,0.74}
\usepackage[pagebackref,breaklinks,colorlinks,citecolor=cvprblue]{hyperref}
\usepackage{amsmath}
\def\paperID{4010} 
\def\confName{CVPR}
\def\confYear{2024}

\title{Hide in Thicket: Generating Imperceptible and Rational Adversarial Perturbations on 3D Point Clouds}

\author{Tianrui Lou$^{1}$ \; Xiaojun Jia$^{2,*}$ \; Jindong Gu$^{3}$ \; 
Li Liu$^{4}$ \; Siyuan Liang$^{5}$ \; Bangyan He$^{6}$ \; Xiaochun Cao$^{1,*}$ \\
\normalsize{$^{1}$Sun Yat-Sen University \quad $^{2}$Nanyang Technological University \quad $^{3}$University of Oxford} \\
\normalsize{$^{4}$National University of Defense Technology \quad $^{5}$National University of Singapore \quad $^{6}$Chinese Academy of Sciences} \\
{\tt\small \{loutianrui, jiaxiaojunqaq, pandaliang521\}@gmail.com \quad jindong.gu@siemens.com} \\
{\tt\small li.liu@oulu.fi \quad hebangyan@iie.ac.cn \quad caoxiaochun@mail.sysu.edu.cn}
}

\begin{document}

\maketitle
\begin{abstract}
Adversarial attack methods based on point manipulation for 3D point cloud classification have revealed the fragility of 3D models, yet the adversarial examples they produce are easily perceived or defended against.
The trade-off between the imperceptibility and adversarial strength leads most point attack methods to inevitably introduce easily detectable outlier points upon a successful attack.
Another promising strategy, shape-based attack, can effectively eliminate outliers, but existing methods often suffer significant reductions in imperceptibility due to irrational deformations.
We find that concealing deformation perturbations in areas insensitive to human eyes can achieve a better trade-off between imperceptibility and adversarial strength, specifically in parts of the object surface that are complex and exhibit drastic curvature changes.
Therefore, we propose a novel shape-based adversarial attack method, HiT-ADV, 
which initially conducts a two-stage search for attack regions based on saliency and imperceptibility scores, 
and then adds deformation perturbations in each attack region using Gaussian kernel functions.
Additionally, HiT-ADV is extendable to physical attack. 
We propose that by employing benign resampling and benign rigid transformations, we can further enhance physical adversarial strength with little sacrifice to imperceptibility.
Extensive experiments have validated the superiority of our method in terms of adversarial and imperceptible properties in both digital and physical spaces. 
Our code is avaliable at: \url{https://github.com/TRLou/HiT-ADV}.

\end{abstract}    
\section{Introduction}
\label{sec:intro}


\begin{figure}[t]
  \centering
   \includegraphics[width=1\linewidth]{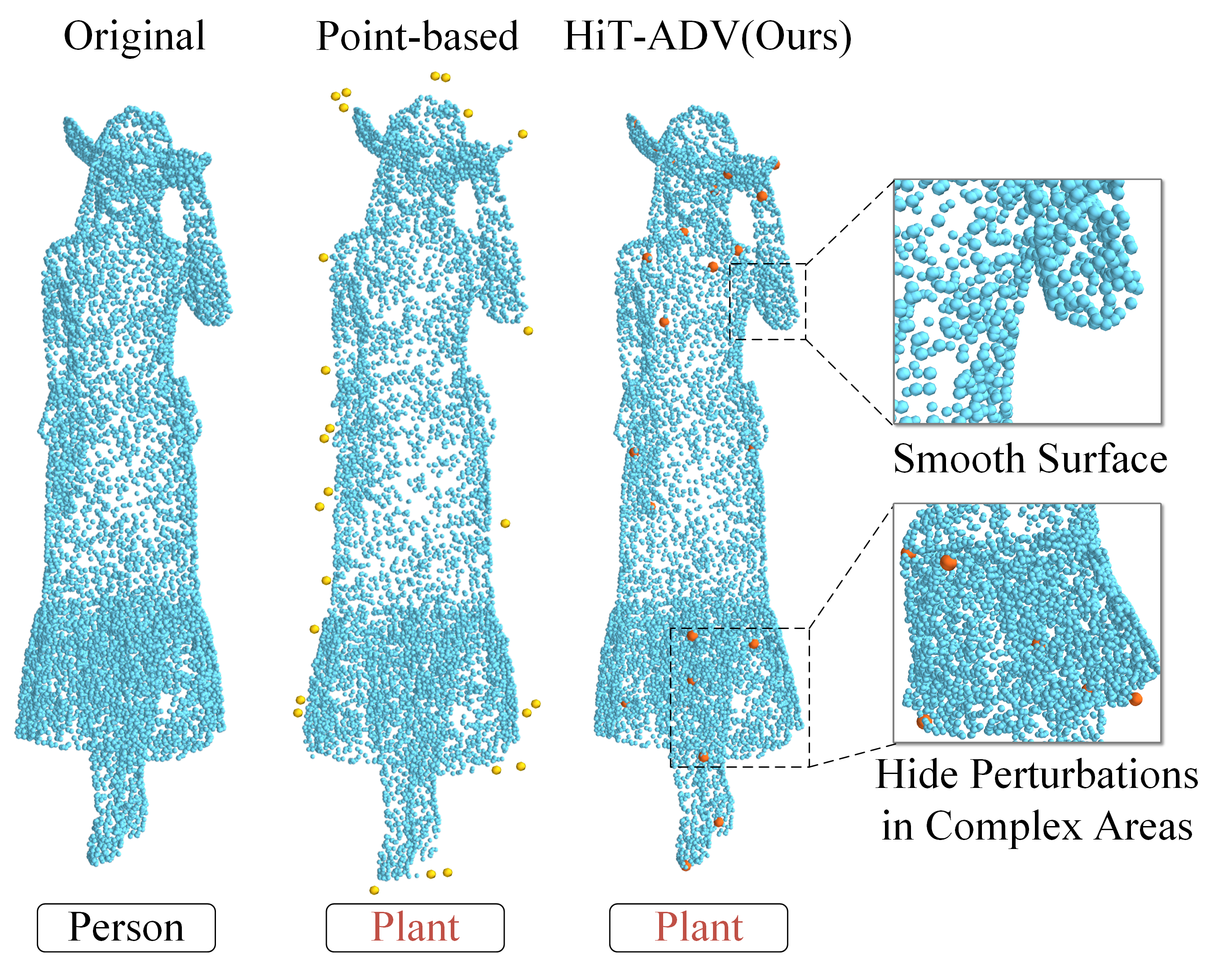}

   \caption{Comparison of successful adversarial examples generated by point-based attack method~\cite{liu2019extending} and HiT-ADV. The orange points represent the central points of local shape deformations generated in HiT-ADV and the yellow points represent the outliers in point-based attack. HiT-ADV has no outlier points and exhibits a smooth surface. By concealing perturbations in complex areas, the deformation perturbations become difficult to perceive.}
   \label{fig:teaser}
   \vspace{-2mm}
\end{figure}

In recent years, numerous models based on Deep Neural Networks (DNNs)~\cite{he2016resnet, simonyan2014vgg, zagoruyko2016wideresnet, gu2021interpretable, vaswani2017transformer} have shown strong performance in a broad spectrum of tasks, including 3D data processing and analysis~\cite{qi2017pointnet, huang2021registration, tang2022segmentation, zhou2018voxelnet, shi2019pointrcnn}.
However, it has been observed that meticulously crafted, imperceptible perturbations can lead DNNs to produce erroneous outputs~\cite{goodfellow2014FGSM, madry2017pgd, jia2019comdefend, gu2022segpgd, jia2020advwatermark}, highlighting their inherent vulnerability. 
This characteristic of DNNs significantly hinders their development in some safety-critical areas, e.g. autonomous driving~\cite{levinson2011autonomousdriving, wang2021lidarattack, tu2020physically, deng2020adanalysis, cao2019lidarad} and robotic task planning~\cite{lin2022manipulation, vemprala2021advrobot}.
Numerous point cloud adversarial attacks have effectively compromised 3D point cloud models~\cite{tsai2020knn, wen2020geoa3, he2023transfer, zhou2020lggan}, exposing their vulnerabilities.
However, adversarial examples in point clouds are usually easily perceived by the human vision compared to those in other domains such as imaging, because point cloud data lacks RGB information to conceal perturbations.
Thus, achieving imperceptibility, a vital attribute of adversarial examples, remains a critical question for further investigation.


Existing adversarial attack methods for point clouds are primarily point-based, encompassing point movement~\cite{liu2019extending, huang2022siadv, wen2020geoa3, tsai2020knn, liu2022aof}, point addition~\cite{xiang2019generating}, and point deletion~\cite{zheng2019saliency}.
However, as shown in Fig.\ref{fig:teaser}, point-based attack methods often result in noticeable outlier points and coarse surfaces.
This directly makes the adversarial examples easily perceptible to the human eyes and prone to be defended by pre-processing defense methods~\cite{zhou2019dupnet, wu2020ifdefense}.
Furthermore, when digital adversarial point clouds are converted into physical adversarial objects, retaining outliers causes noticeable shape distortion, while discarding them leads to a significant decrease in adversarial strength.
While many prior studies have considered geometric features, using regularization terms to constrain optimization objectives in order to reduce outlier points 
~\cite{wen2020geoa3, tsai2020knn, xiang2019generating}, these methods face challenges in fundamentally eliminating them 
due to the trade-off between imperceptibility and adversarial strength. 
Another attack strategy is based on deformation~\cite{liu2020sink, dong2022isometric, zhang2023meshattack, tang2023manifold}, which can effectively prevent outlier generation by reinforcing inter-point correlations. However, current deformation attack methods typically induce excessive and irrational perturbations, extremely compromising the imperceptibility of successful adversarial examples.

In fact, we find that the reason previous works have difficulty optimizing a satisfactory trade-off relationship is due to a misconception about imperceptibility. 
We argue that imperceptibility can be maintained with reasonable shape changes, not just by 
keeping
the original shape exactly.
Specifically, we find that the human eyes have varying sensitivity to different regions of an object. 
Concretely, they can confidently discern areas that should be smooth, but it is challenging to distinguish whether inherently complex surfaces have been tampered with.
Therefore, concealing adversarial perturbations in complex surfaces can maintain the shape reasonable and the perturbations imperceptible.

Based on this observation, we propose a novel imperceptible shape-based attack, called HiT-ADV.
Firstly, we introduce a two-stage attack region search module to search multiple regions that are vulnerable to attack and insensitive to the human vision.
Secondly, we conduct an iterative attack using Gaussian kernel functions to calculate deformation perturbations. 
Correspondingly, we propose novel imperceptibility regularization terms to constrain deformation perturbations, continuously refining the shape of adversarial point clouds to ensure their imperceptibility.
Finally, we analyze the main challenge in extending to physical attack, which is the vanishing of digital adversarial strength during the physical attack process.
The vanished digital adversarial strength primarily comes from outlier points, non-shape-altering point movements, and adversarial rigid transformations.
To tackle this issue, we perform benign resampling and benign rigid transformations before every iteration to ensure our adversarial properties are mainly from physical deformation.
Through this way, HiT-ADV can be effectively extended to physical attack.
Comprehensive experiments on two benchmarks with classifiers in two representative DNN architectures validate that HiT-ADV is imperceptible and effective, and outperforms the state-of-the-art methods.
Additionally, we validate the effectiveness of HiT-ADV in physical scenarios and visualize the attack results.


Our main contributions are in three aspects:
\begin{itemize}
\item We propose an imperceptible shape-based attack framework. By concealing shape perturbations in complex surface areas, we achieve a favorable trade-off between adversarial strength and imperceptibility.

\item We propose an optimization method that effectively suppresses the digital adversarial strength of adversarial examples, enhancing their performance in the physical world.

\item 	We validate the superiority of our framework to the state-of-the-art point cloud attack methods via extensive experiments. 
Besides, we print the adversarial examples generated by HiT-ADV and use the re-scanned point clouds to successfully attack the classification model, to verify the feasibility of our method when extended to the physical world.

\end{itemize}

\section{Related Work}
\label{sec:related}
\paragraph{DNN Models for 3D Classification.}
Point cloud is a promising method of 3D data representation, effectively capturing the geometric information of real-world objects, and can be extensively acquired through LiDAR or RGB-D cameras. 
However, their irregular and unordered nature poses challenges for direct utilization.
Early research efforts attempted to transform them into structured formats, such as voxels, multi-view images, or meshes, for subsequent tasks.
Until recently, PointNet~\cite{qi2017pointnet} innovatively tackled the irregularity and unordered nature of point cloud data by employing multi-layer perceptrons (MLP) and max-pooling layer, facilitating end-to-end classification and segmentation directly on point clouds.
Due to the convenience and high performance of PointNet, numerous research efforts were inspired to propose various DNN architectures for direct point cloud learning. 
Examples include PointNet++~\cite{qi2017pointnet++}, which utilizes MLPs to learn hierarchical point cloud features, DGCNN~\cite{wang2019dgcnn} with graph neural networks, PCT~\cite{guo2021pct} with the transformer architecture, and so forth~\cite{wu2019pointconv, thomas2019kp, li2018sonet, liu2019rscnn}.

\paragraph{3D Adversarial Attack.}
The vulnerability of 3D DNN models mirrors that of the 2D image models~\cite{goodfellow2014FGSM}, where a slight perturbation, imperceptible to the human eye, can easily lead to erroneous model classification.
Liu et al.~\cite{liu2019extending} first extended the gradient-based adversarial attack, FGM~\cite{goodfellow2014FGSM}, to the point cloud domain under an $l_2$ norm constraint. 
Xiang et al.~\cite{xiang2019generating} introduced the C\&W attack framework~\cite{carlini2017cw} in point cloud attacks, producing superior adversarial examples by shifting point coordinates. 
Moreover, they proposed achieving adversarial properties by adding points, clusters, and objects. 
Besides, Zheng et al.~\cite{zheng2019saliency} proposed that deleting a small number of points with high saliency can effectively cause misclassification in point cloud classification models.

Since point clouds only contain coordinate information and lack the RGB three-channel data of images to conceal adversarial perturbations, coupled with the human eye's high sensitivity to outliers in adversarial point clouds, many researchers have been dedicated to making adversarial perturbations more imperceptible. 
Tsai et al.~\cite{tsai2020knn} improved the C\&W attack framework by introducing a KNN regularization term, resulting in a tighter point cloud surface and suppressing the generation of outlier points. 
Wen et al.~\cite{wen2020geoa3} introduced geometry-aware regularization constraints, which encompass maintaining local curvature consistency and ensuring a uniform surface on the adversarial point cloud.
Huang et al.~\cite{huang2022siadv} first transformed the coordinate system onto the tangent plane and then introduced perturbations on this plane to generate shape-invariant point clouds.

These point-based methods above 
enhanced imperceptibility  through geometric perspectives while struggling to  eliminate all outliers and obtain smooth surfaces.
Thus, many studies have attempted to propose shape-based attack methods.
Zhang et al.~\cite{zhang2023meshattack} and Miao et al.~\cite{dong2022isometric} suggested directly attacking mesh data to generate sufficiently smooth results, utilizing edge length regularization and Gaussian curvature regularization respectively.
Liu et al.~\cite{liu2020sink} attempted to conduct shape-based attacks by adding new features to objects, like adversarial sticks. Additionally, they proposed using the Gaussian kernel for sinking operations to achieve point cloud shape deformation.
Tang et al.~\cite{tang2023manifold} proposed to adversarially stretch the latent variables in an auto-encoder, which can be decoded as smooth adversarial point clouds.
While these shape-based methods eliminate outlier points and ensure smoothness, they either face challenges in data acquisition or are easily perceptible. 
In contrast, our proposed shape-based attack HiT-ADV is both smooth and imperceptible, and these characteristics enable it to be easily extended to the physical scenarios.

\section{Preliminaries}
\label{sec:preli}
We focus on adversarial attacks in the context of point cloud classification tasks. 
We denote $\mathcal{D} = \{\mathcal{P}_i, y_i\}^n_{i=1}$ as the training dataset of a group of point clouds.
We use $\mathcal{P}_i \in \mathbb{R}^{m \times 3}$ to represent a single point cloud to be classified in $\mathcal{D}$, which contains a set of unordered point sets, denoted as $\{\boldsymbol{p}_j\}$, and $y_i$ is the corresponding label. Each point $\boldsymbol{p}_j \in \mathbb{R}^3$ in this set contains three-dimensional coordinate information.
The goal of the adversarial attack is to introduce a small perturbation $\boldsymbol{\delta}_i$ to the point cloud $\mathcal{P}_i$, such that the classifier $\mathcal{F}$ makes an incorrect classification of $\mathcal{P}_i$, i.e. $\mathcal{F}(\mathcal{P}_i + \boldsymbol{\delta}_i) \neq y_i$.
Note that the situation discussed in this paper is assumed to be untargeted adversarial attack by default. 
For the visualization results of targeted attack, please refer to the appendix.

In order to obtain more imperceptible adversarial point clouds $\mathcal{P}' = \mathcal{P} + \boldsymbol{\delta}$, our attack process adopts a gradient-based optimization algorithm, following ~\cite{xiang2019generating}, and can be formulated as:
\begin{equation}
\begin{aligned}
\min \; \mathcal{L}_{cls}(\mathcal{P}') + \lambda * \mathcal{L}_{dis}(\mathcal{P}, \mathcal{P'}),
\label{eq:cw}
\end{aligned}
\end{equation}
where $\mathcal{L}_{cls}$ is an adversarial loss function measuring the successful attack probability and $\mathcal{L}_{dis}$ is a regularization term constraining the distance between benign point cloud $\mathcal{P}$ and the adversarial one $\mathcal{P}'$.
$\lambda$ is a hyper-parameter, automatically optimized to an approximate optimal value through binary search, which can balance the trade-off relationship between the two terms above.
Concretely, $\mathcal{L}_{cls}$ can be formulated as:

\begin{equation}
\begin{aligned}
\mathcal{L}_{cls} = \max \; \{ -\kappa, \mathcal{Z}_t(\mathcal{P}') - \underset{j \neq t}{\max} \mathcal{Z}_j(\mathcal{P}') \},
\label{eq:clsloss}
\end{aligned}
\end{equation}
where $\kappa\geq0$ represents a margin threshold, and $\mathcal{Z}$ represents the output logits of $\mathcal{F}(\mathcal{P}')$, and $t$ is the true label class.

\paragraph{Problem Analysis.}

The point-based adversarial perturbation $\boldsymbol{\delta} \in \mathbb{R}^3$ is calculated from the back-propagated gradient of the loss in Eq.\ref{eq:cw}.
Even though some methods, such as~\cite{xiang2019generating, tsai2020knn, wen2020geoa3, dong2022isometric}, attempt to eliminate outliers and ensure overall imperceptibility by modifying the regularization terms $\mathcal{L}_{dis}$, due to the lack of hard correlation between the $\boldsymbol{\delta}_j$ of different points, there will always be a small number of points escaping the regularization term constraint when the optimization objective $\mathcal{L}_{cls}$ becomes difficult.
Please refer to the visualization result of Fig.\ref{fig:vis}.
Furthermore, excessively demanding the shape of adversarial point clouds to mirror the original, while also lacking outlier points, severely restricts the search space, leading to poor adversarial strength.
As compared to point-based attack methods, shape-based attacks, such as \cite{liu2020sink}, often achieve smooth surfaces and eliminate outliers because they rigidly constrain the relationship between the perturbations $\boldsymbol{\delta}_j$ of different points.
However, shape-based attacks often require a larger search space. 
Therefore, ensuring the imperceptibility of shape-based attacks remains a critical challenge that requires resolution.
We have an intuition that the human eyes have different sensitivities to changes in different parts of a 3D object. 
Deformations occurring in areas with complex surfaces are less likely to be perceived by human eyes because people do not have an absolute prior cognition of the specific shape of these areas.
Based on this intuition, we propose an imperceptible shape-based attack method HiT-ADV, which will be introduced in detail in the next section.
\section{Methodology}

\begin{figure*}[t]
  \centering
   \includegraphics[width=1\linewidth]{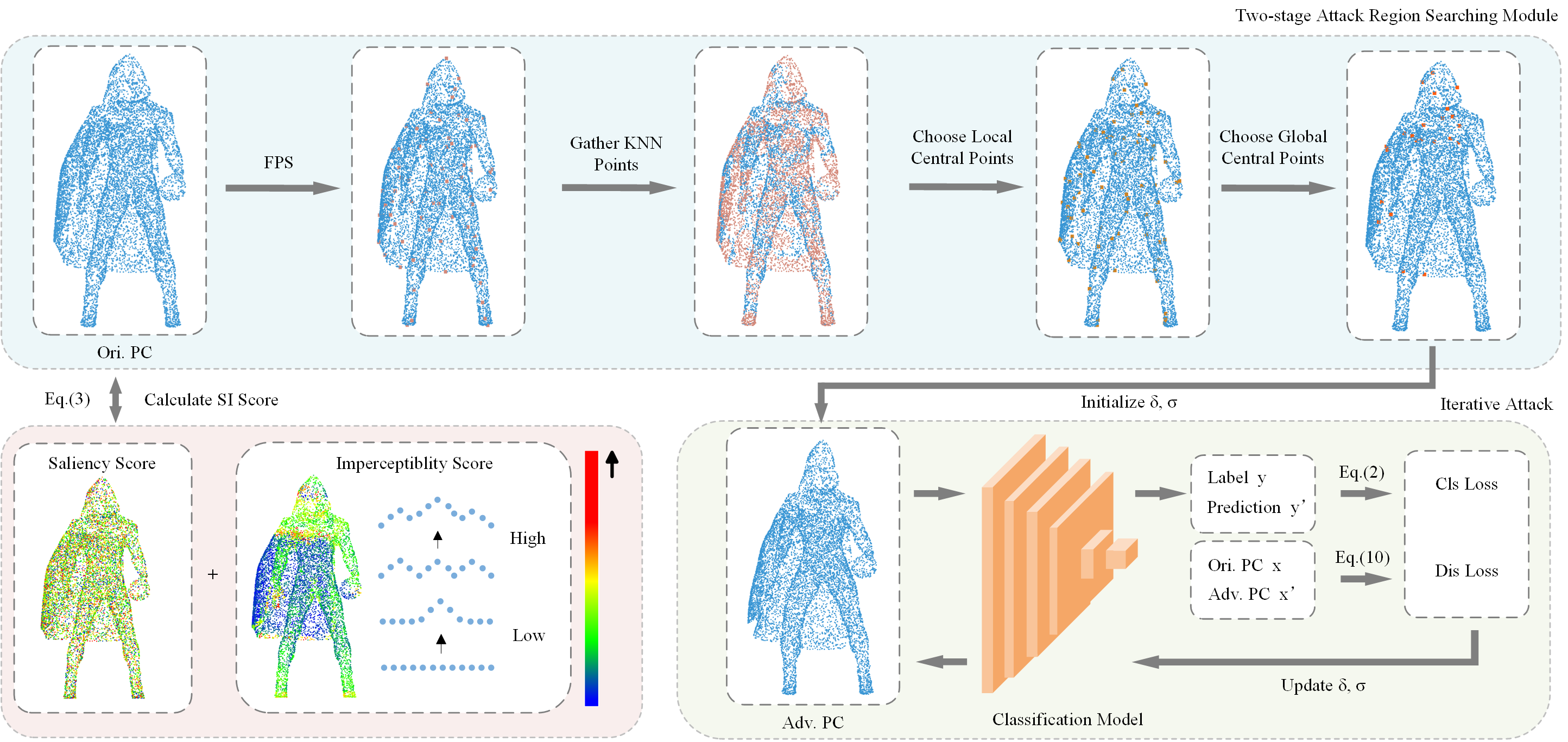}
   \caption{Demonstration of the framework of HiT-ADV. For clean point cloud samples, HiT-ADV first calculates the SI score for each point, and we color them in this figure according to their ranking. Subsequently, we employ a two-stage attack region search to locate the global central points of imperceptible regions. Finally, we iteratively attack using multiple Gaussian kernel functions and propose a new distance loss for constraint.}
   \label{fig:framework}
   \vspace{-1mm}
\end{figure*}

In this section, we introduce the specific framework and implementation of HiT-ADV, as well as methods to further suppress digital adversarial strength. 
Please refer to Fig.\ref{fig:framework} for the demonstration of our framework.

\subsection{Saliency and Imperceptibility Score}
The design objective of this module is to locate areas in the object that are both likely to affect the classification result and imperceptible to human vision. 
Note that when a region undergoes deformation, the center of this region often suffers the most drastic perturbations.
Therefore, we can effectively search suitable central points for deformation perturbations in the clean point cloud $\mathcal{P}_i$ using a saliency and imperceptibility score (SI score) $\mathcal{S}$.
The definition of saliency score $\mathcal{S}_1$ follows~\cite{zheng2019saliency}, which measures the contribution of each point to the correct classification. 
As discussed in Section~\ref{sec:preli}, points located in complex surface areas should be evaluated with higher imperceptibility scores $\mathcal{S}_2$.
Therefore, we propose to formulate $\mathcal{S}_2$ as follow:

\begin{equation}
\vspace{-1mm}
\begin{aligned}
\mathcal{S}_2(\boldsymbol{p}_j;\mathcal{P}) &= \sqrt{\frac{1}{k}\sum\limits_{\boldsymbol{q} \in \mathcal{N}_{\boldsymbol{p}_j}}(\mathcal{C}(\boldsymbol{q};\mathcal{P}) - \mu)^2}, \\
\mu &= \frac{1}{k}\sum\limits_{\boldsymbol{q} \in \mathcal{N}_{\boldsymbol{p}_j}}\mathcal{C}(\boldsymbol{q};\mathcal{P}),
\label{eq:S2}
\end{aligned}
\end{equation}
where $\mathcal{N}_{\boldsymbol{p}_j}$ represents the k nearest neighbor of point $\boldsymbol{p}_j$. 
And $\mathcal{C}$ is the local curvature of $\boldsymbol{p}_j$ whose formulation follows~\cite{wen2020geoa3}:
\begin{equation}
\begin{aligned}
\mathcal{C}(\boldsymbol{p}_j;\mathcal{P}) = \frac{1}{k}\sum\limits_{\boldsymbol{q} \in \mathcal{N}_{\boldsymbol{p}_j}}|<(\boldsymbol{q} - \boldsymbol{p}_j)/||\boldsymbol{q} - \boldsymbol{p}_j||_2, \boldsymbol{n}_{\boldsymbol{p}_j}>|,
\label{eq:curv}
\end{aligned}
\end{equation}
where $\boldsymbol{n}_{\boldsymbol{p}_j}$ denotes the normal vector of the local surface around $\boldsymbol{p}_j$, and $<\cdot, \cdot>$ denotes inner product. 
$\mathcal{S}_2$ evaluates the standard deviation of local surface curvature, describing the intensity of local curvature variation.
We can normalize $\mathcal{S}_1$ and $\mathcal{S}_2$ to the range $[0, 1]$ and add them together to obtain the SI score $\mathcal{S} = \mathcal{S}_1 + \alpha \cdot \mathcal{S}_2$, where $\alpha$ is set to $1$ by default.

\subsection{Two-stage Attack Region Searching Module}
To perform adversarial deformation on different local regions, we first partition the point cloud $\boldsymbol{p}_j$.
Concretely, we obtain $n$ points by farthest point sampling (FPS) and gather $k$ nearest neighbor points around them.
From these regions, we select the points with the highest SI scores as their central points.
To further enhance imperceptibility, we design the second-stage selection to choose the final deformation regions, by selecting $\tilde{n}$ central points, denoted as $\{\tilde{\boldsymbol{p}}_i\}_{i=1}^{\tilde{n}}$, with the highest SI scores from the $n$ candidate points in the previous stage.
Through this mechanism, we ensure that the deformation perturbations can be concealed in complex surface areas.

\subsection{Shape-based Attack Method based on Gaussian Kernel Function}
For the obtained central points, we add random perturbations $\boldsymbol{\delta}$ to them as initialization.
Then we need to diffuse the perturbations, calculating the size and direction of perturbations for each point around the central points.
To make the adversarial point cloud surface smooth, we use the classical Gaussian kernel function to calculate the adversarial perturbations.
Given a central point $\tilde{\boldsymbol{p}}$ and an original point $\boldsymbol{p}$, we can formulate the Gaussian kernel function as:
\begin{equation}
\begin{aligned}
Gauss(\tilde{\boldsymbol{p}}, \boldsymbol{p}) = e^{-\frac{||\tilde{\boldsymbol{p}}-\boldsymbol{p}||_2^2}{2\sigma^2}},
\label{eq:gauss}
\end{aligned}
\end{equation}
where $\boldsymbol{\sigma}$ is the hyper-parameter of bandwidth controlling the shape of the Gaussian kernel function. 
Note that the magnitude and direction of the final perturbation of an individual point should be collectively calculated by Gaussian kernel functions of multiple different deformation central points.
Thus we use Nadaraya-Watson kernel regression~\cite{watson1964smooth} to calculate the adversarial point $\boldsymbol{p}_j'$:
\begin{equation}
\begin{aligned}
\boldsymbol{p}_j' = \frac{\sum^{\tilde{n}}_{i=1}(\boldsymbol{p}_j + \boldsymbol{\delta}_i) \cdot Gauss(\tilde{\boldsymbol{p}}_i, \boldsymbol{p}_j)}{\sum^{\tilde{n}}_{i=1} Gauss(\tilde{\boldsymbol{p}}_i, \boldsymbol{p}_j)}
\label{eq:regression}
\end{aligned}
\end{equation}
Both $\boldsymbol{\delta}$ and $\boldsymbol{\sigma}$ can affect the generation of the adversarial point cloud, so their values will be continuously updated during the iterative optimization of the attack process.

To constrain the imperceptibility of the adversarial examples, we use three regularization terms. 
The first one is denoted as kernel loss $\mathcal{L}_{ker}$, formulated as:
\begin{equation}
\begin{aligned}
\mathcal{L}_{ker} = ||\boldsymbol{\delta}||_2 + ||a-\boldsymbol{\sigma}||_2, 
\label{eq:kernelloss}
\end{aligned}
\end{equation}
where $a$ is a hyper-parameter and $\boldsymbol{\sigma}$ is clipped to ensure it is smaller than $a$.
$\mathcal{L}_{ker}$ ~requires the perturbations on central points to be as small as possible, and the Gaussian kernel function to be as flat as possible to prevent drastic local spikes. 
Further, we fine-tune the deformation perturbations based on the surface complexity, allowing more severe perturbations to be concealed in more complex areas.
The second regularization term describes the relationship between the $\boldsymbol{\sigma}$ of the Gaussian kernel and the standard deviation of curvature at the central points $\tilde{\boldsymbol{p}}$ by cosine similarity:
\begin{equation}
\begin{aligned}
\mathcal{L}_{hide} = \frac{\boldsymbol{\sigma} \cdot \mathcal{C}_{std}(\tilde{\boldsymbol{p}}_i;\mathcal{P})}{||\boldsymbol{\sigma}||_2 \times ||\mathcal{C}_{std}(\tilde{\boldsymbol{p}}_i;\mathcal{P})||_2},
\label{eq:hideloss}
\end{aligned}
\end{equation}
where $\mathcal{C}_{std}(\tilde{\boldsymbol{p}}_i;\mathcal{P})$ is equally defined as $\mathcal{S}_2$, and both $\mathcal{C}_{std}$ and $\boldsymbol{\sigma}$ are normalized to the range [0, 1].
This regularization term loss results in smaller $\boldsymbol{\sigma}$ values for the Gaussian kernels at deformation center points located on complex surfaces and larger $\boldsymbol{\sigma}$ values for smoother surfaces.
The final regularization term is the typical Chamfer loss $\mathcal{L}_{cha}$:
\begin{equation}
\begin{aligned}
\mathcal{L}_{cha} &= \frac{1}{m} \sum\limits_{\boldsymbol{p} \in \mathcal{P}} \min_{\boldsymbol{p}' \in \mathcal{P}'} ||\boldsymbol{p} - \boldsymbol{p}'||^2_2 \\&+ \frac{1}{m'} \sum\limits_{\boldsymbol{p}' \in \mathcal{P}'} \min_{\boldsymbol{p} \in \mathcal{P}} ||\boldsymbol{p}' - \boldsymbol{p}||^2_2
\label{eq:chamferloss}
\end{aligned}
\end{equation}
$\mathcal{L}_{dis}$ in Eq.\ref{eq:cw} can be expressed with 3 hyper-parameters as:
\begin{equation}
\begin{aligned}
\mathcal{L}_{dis} = \lambda_1 \mathcal{L}_{ker} + \lambda_2 \mathcal{L}_{hide} + \lambda_3 \mathcal{L}_{cha},
\label{eq:disloss}
\end{aligned}
\end{equation}

\subsection{Suppressing Digital Adversarial Strength}
\label{sec:physical}
To further enhance the physical adversarial strength of the generated adversarial point clouds by HiT-ADV, we should avoid adversarial strength obtained by rigid transformations and non-shape-altering point movements.
Thus before each iteration of attack, we need to minimize the adversarial strength above as much as possible.
Concretely, we use the MaxOT~\cite{dong2022isometric} optimization paradigm to suppress adversarial strength from rigid transformations:
\begin{equation}
\begin{aligned}
\boldsymbol{\mathcal{\tilde{T}}} = \text{argmax}_{\boldsymbol{\mathcal{T}}} \; \mathcal{L}_{cls}( \boldsymbol{\mathcal{T}}(\mathcal{P}'), y)
\label{eq:maxot}
\end{aligned}
\end{equation}

The rigid transformation can be formulated as:
\begin{equation}
\begin{aligned}
\boldsymbol{\mathcal{T}}(\mathcal{P}') = \boldsymbol{S}(\boldsymbol{R}(\mathcal{P}')) + \boldsymbol{T},
\label{eq:rigidtransformation}
\end{aligned}
\end{equation}
where $\boldsymbol{S}, \boldsymbol{R}, \boldsymbol{T}$ are scaling, rotation and translation matrices respectively.
Note that the benign resampling process is non-differentiable and cannot be directly incorporated into the MaxOT optimization paradigm. 
Therefore, we choose to upsample before rigid transformation and then use farthest point sampling to downsample and match the number of points with the original $\mathcal{P}$ after the iteration.
Through these two steps, we can effectively suppress digital adversarial strength during the iteration process and the overall optimization objective is reformulated as:
\begin{equation}
\begin{aligned}
\min \; \mathcal{L}_{cls}(\boldsymbol{\mathcal{\tilde{T}}}(\mathcal{P}')) + \lambda * \mathcal{L}_{dis}(\boldsymbol{\mathcal{\tilde{T}}}(\mathcal{P}), \boldsymbol{\mathcal{\tilde{T}}}(\mathcal{P'}))
\label{eq:rigidtransformation}
\end{aligned}
\end{equation}

\begin{table*}[t]
\centering
\scalebox{0.8}{
\begin{tabular}{@{}c|c|cccc|cccc@{}}
\toprule
\multicolumn{1}{l|}{}      &                                                             & \multicolumn{4}{c|}{ModelNet40}                                                                                                           & \multicolumn{4}{c}{ShapeNet Part}                                                                                                         \\ \midrule
Model                      & Method                                                      & ASR(\%) $\uparrow$               & CSD $\downarrow$                 & Uniform $\downarrow$             & KNN(*1e-3) $\downarrow$          & ASR(\%) $\uparrow$               & CSD $\downarrow$                 & Uniform $\downarrow$             & KNN(*1e-3) $\downarrow$          \\ \midrule
\multirow{10}{*}{\rotatebox{90}{PointNet}} & IFGSM($l_{inf}$)                                            & 99.01                            & 2.6341                           & 0.3134                           & 0.8558                           & 95.20                            & 2.8831                           & 0.3035                           & 0.8076                           \\
                           & IFGM($l_2$)~\cite{liu2019extending}   & 99.68                            & 1.6502                           & 0.3173                           & 0.7886                           & 97.49                            & 2.3401                           & 0.2404                           & 0.5898                           \\
                           & 3D-ADV~\cite{xiang2019generating}     & \textbf{100.00} & 0.9500                           & 0.2965                           & 0.7355                           & \textbf{100.00} & 2.1603                           & 0.2103                           & 0.5322                           \\
                           & KNN~\cite{tsai2020knn}                & 99.68                            & 2.3133                           & 0.3695                           & \textbf{0.6077} & 96.40                            & 2.5067                           & 0.2723                           & \textbf{0.5098} \\
                           & GeoA$^3$~\cite{wen2020geoa3}          & \textbf{100.00} & 1.7112                           & 0.2919                           & 0.7174                           & \textbf{100.00} & 3.7291                           & 0.2658                           & 0.6892                           \\
                           & SI-ADV~\cite{huang2022siadv}          & 99.32                            & 1.4601                           & 0.3059                           & 1.5979                           & 96.38                            & 2.8476                           & 0.2945                           & 0.9217                           \\
                           & AOF~\cite{liu2022aof}                 & \textbf{100.00} & 2.8213                           & 0.3809                           & 1.2519                           & 99.85                            & 3.3172                           & 0.3601                           & 1.3110                           \\
                           & MeshAttack~\cite{zhang2023meshattack} & 96.64                            & 2.2161                           & 0.3008                           & 0.7971                           & 97.24                            & 3.0671                           & 0.3444                           & 0.7798                           \\
                           & Eps-iso~\cite{dong2022isometric}      & 97.17                            & 2.2514                           & 0.3053                           & 0.7949                           & 98.01                            & 2.6093                           & 0.2908                           & 0.7457                           \\
                           & HiT-ADV (Ours)                                              & \textbf{100.00} & \textbf{0.4709} & \textbf{0.2883} & 0.7447                           & \textbf{100.00} & \textbf{0.9810} & \textbf{0.1874} & 0.5272                           \\ \midrule
\multirow{10}{*}{\rotatebox{90}{DGCNN}}    & IFGSM($l_{inf}$)                                            & 98.96                            & 2.7243                           & 0.3112                           & 0.8891                           & 98.46                            & 3.3550                           & 0.2094                           & 0.7085                           \\
                           & IFGM($l_2$)~\cite{liu2019extending}   & 98.96                            & 2.0193                           & 0.2983                           & 0.6846                           & 99.51                            & 2.6947                           & 0.2080                           & 0.6627                           \\
                           & 3D-ADV~\cite{xiang2019generating}     & \textbf{100.00} & 1.0206                           & 0.2919                           & 0.8677                           & \textbf{100.00} & 2.3788                           & 0.1886                           & 0.6837                           \\
                           & KNN~\cite{tsai2020knn}                & 96.15                            & 3.2322                           & 0.3987                           & \textbf{0.6080} & 98.09                            & 3.4718                           & 0.2372                           & 0.6349                           \\
                           & GeoA$^3$~\cite{wen2020geoa3}          & 99.71                            & 1.0286                           & 0.2887                           & 0.8651                           & \textbf{100.00} & 2.9907                           & 0.1895                           & 0.7180                           \\
                           & SI-ADV~\cite{huang2022siadv}          & 96.08                            & 2.3211                           & 0.5557                           & 1.5992                           & 96.43                            & 3.0193                           & 0.2456                           & 0.6824                           \\
                           & AOF~\cite{liu2022aof}                 & 97.09                            & 2.3148                           & 0.3226                           & 1.0211                           & 98.54                            & 3.2227                           & 0.2444                           & 0.8970                           \\
                           & MeshAttack~\cite{zhang2023meshattack} & \textbf{100.00} & 1.0411                           & 0.3192                           & 0.8366                           & 99.43                            & 3.1836                           & 0.2232                           & 0.7092                           \\
                           & Eps-iso~\cite{dong2022isometric}      & \textbf{100.00} & 1.0430                           & 0.3001                           & 0.8377                           & \textbf{100.00} & 3.0629                           & 0.2171                           & 0.7108                           \\
                           & HiT-ADV (Ours)                                              & \textbf{100.00} & \textbf{0.1946} & \textbf{0.2867} & 0.8374                           & \textbf{100.00} & \textbf{0.6629} & \textbf{0.1870} & \textbf{0.6325} \\ \bottomrule
\end{tabular}
}
\caption{Comparison of different point cloud attack methods on the ModelNet40 and ShapeNet datasets targeting an undefended PointNet and DGCNN classification model, including adversarial attack success rate (ASR), curvature standard deviation difference (CSD), uniform metric, and KNN distance.
}
\label{tab:main}
\end{table*}

\section{Experiments}

\begin{table*}[]
\centering
\scalebox{0.8}{
\begin{tabular}{@{}c|ccccc|ccccc@{}}
\toprule
               & \multicolumn{5}{c|}{PointNet}                                                                                                                                            & \multicolumn{5}{c}{DGCNN}                                                                                                                                                 \\ \midrule
Method         & No Defense                       & SRS                             & SOR                             & DUP-Net                         & AT                              & No Defense                       & SRS                             & SOR                             & DUP-Net                         & AT                               \\ \midrule
IFGSM          & 99.01                            & 77.37                           & 60.68                           & 43.57                           & 49.09                           & 98.96                            & 67.31                           & 82.14                           & 61.10                           & 72.06                            \\
IFGM~\cite{liu2019extending}           & 99.68                            & 81.81                           & 33.49                           & 21.49                           & 21.65                           & 98.96                            & 58.62                           & 72.54                           & 59.77                           & 42.58                            \\
3D-ADV~\cite{xiang2019generating}         & \textbf{100.00} & 24.28                           & 24.25                           & 14.03                            & 60.98                           & \textbf{100.00} & 35.55                           & 43.08                           & 52.33                           & 74.82                            \\
KNN~\cite{tsai2020knn}            & 99.68                            & 84.72                           & 28.21                           & 21.27                           & 11.39                           & 96.15                            & 74.26                           & 79.88                           & 66.25                           & 52.41                            \\
GeoA3~\cite{wen2020geoa3}          & \textbf{100.00} & 40.66                           & 18.76                           & 13.65                           & 64.35                           & 99.71                            & \textbf{76.28} & 44.61                           & 53.00                           & 77.30                            \\
SI-ADV~\cite{huang2022siadv}         & 99.32                            & 58.32                           & 50.20                           & 56.73                           & 36.21                           & 96.08                            & 72.17                           & 71.78                           & 73.02                           & 51.99                            \\
AOF~\cite{liu2022aof}            & \textbf{100.00} & 73.82                           & 86.20                           & 63.27                           & 27.20                           & 97.09                            & 50.81                           & 63.58                           & 60.24                           & 33.62                            \\
MeshAttack~\cite{zhang2023meshattack}     & 96.64                            & 25.67                           & 18.01                           & 18.46                           & 24.21                           & \textbf{100.00} & 32.12                           & 45.66                           & 53.28                           & 74.29                            \\
Eps-iso~\cite{dong2022isometric}        & 97.17                            & 26.16                           & 19.79                           & 19.61                           & 30.23                           & \textbf{100.00} & 31.36                           & 47.28                           & 54.33                           & 78.96                            \\
HiT-ADV (Ours) & \textbf{100.00} & \textbf{84.95} & \textbf{87.86} & \textbf{72.27} & \textbf{99.76} & \textbf{100.00} & 66.82                           & \textbf{83.98} & \textbf{92.37} & \textbf{100.00} \\ \bottomrule
\end{tabular}}
\caption{Comparison results of ASR (\%) for different attack methods on models with and without applied defense methods.
}
\vspace{-3mm}
\label{tab:defend}
\end{table*}

\subsection{Implementation}
We implement our framework and reproduce all the DNN models with PyTorch, and report the results on a workstation with an Intel Xeon Gold 6226R CPU@2.90Hz and 64 GB of memory using a single RTX 3090 GPU. 

\subsection{Experimental Setup}
\paragraph{Datasets.} Our experiment utilize two public point cloud datasets, ModelNet40~\cite{wu2015modelnet} and ShapeNet Part~\cite{chang2015shapenet}, to demonstrate the superiority of HiT-ADV.
The ModelNet40 dataset comprises 12,311 CAD models across 40 primary object classes, split into 9,843 objects for training and 2,468 objects for testing. The ShapeNet Part dataset includes 16,881 pre-aligned shapes from 16 categories, with 12,137 objects designated for training and 2,874 objects for testing.

\paragraph{Point Cloud Classifiers.} We select the two most representative point cloud classification models for experimentation: the MLP-based PointNet~\cite{qi2017pointnet} and the graph-based DGCNN~\cite{wang2019dgcnn}. Before starting the following experiments, we train both classification models using clean examples according to their original papers.

\paragraph{Attack Methods for Comparison.} We select nine state-of-the-art point cloud adversarial attack methods as our baseline for comparison, including IFGSM($l_{inf}$), IFGM($l_2$)~\cite{liu2019extending}, 3D-ADV~\cite{xiang2019generating}, KNN~\cite{tsai2020knn}, GeoA$^3$~\cite{wen2020geoa3}, SI-ADV~\cite{huang2022siadv}, AOF~\cite{liu2022aof}, MeshAttack~\cite{zhang2023meshattack} and Eps-iso~\cite{dong2022isometric}. 

\paragraph{Evaluation Metric.} We assess the efficiency of adversarial attacks through the attack success rate (ASR), which represents the percentage of generated adversarial examples that can deceive the classifiers. 
Additionally, we evaluate the imperceptibility of these attack methods by three distance metrics, including one fully related metric curvature standard deviation distance (CSD), and two no related metrics, i.e. uniform metric~\cite{li2019pugan} and KNN distance~\cite{tsai2020knn}.
The CSD metric intuitively compares the consistency of $\mathcal{C}_{std}$ between clean point clouds and adversarial point clouds. 
Besides providing a macro-level comparison of the similarity, it also evaluates whether the attack methods can generate perturbations reasonably based on the surface complexity. 
CSD can be formulated as:
\begin{equation}
\begin{aligned}
CSD = ||\mathcal{C}_{std}(\mathcal{P}) - \mathcal{C}_{std}(\mathcal{P}’)||_2
\label{eq:CSD}
\end{aligned}
\end{equation}

\begin{figure*}[t]
  \centering
   \includegraphics[width=1\linewidth]{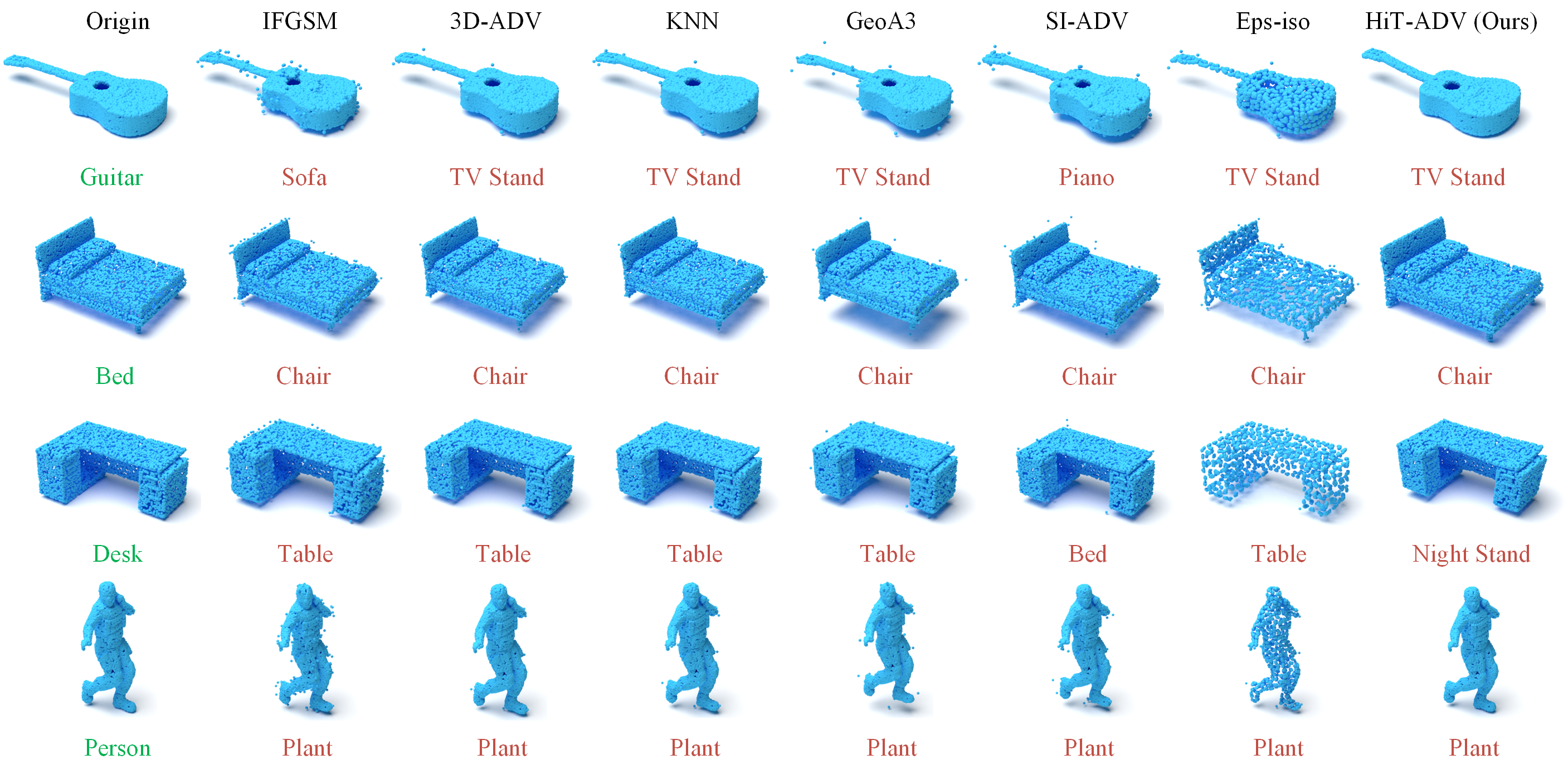}
   \caption{Visualization of original and adversarial point clouds generated by different adversarial attack methods for attacking PointNet.}
   \label{fig:vis}
\end{figure*}

\subsection{Performance Analysis}
\paragraph{Quantitative Results on Imperceptibility.} In order to fairly compare the imperceptibility of various point cloud adversarial attack methods, we adjust the settings of all methods so that they could achieve attack success rates of over 96\% on classification models without defense. 
Under these settings, we compare their three imperceptibility metrics, and the results are shown in Tab.\ref{tab:main}. 
The results show that HiT-ADV can achieve superior performance in CSD and Uniform, while maintaining 100\% attack success rates across all models and datasets. 
Although HiT-ADV gets a bit lower KNN metrics compared to methods that use only KNN distance regularization, its performance is still comparable to other point cloud methods.
It demonstrates that HiT-ADV can generate high-quality adversarial point clouds while achieving better imperceptibility by hiding the perturbations in areas that are insensitive to human vision.
\vspace{-5mm}

\paragraph{Attack Performance against Defense.}
To evaluate the adversarial strength of our proposed deformation attack method, we conducted tests on classification models with four different defense methods, i.e. SRS, SOR, DUP-Net~\cite{zhou2019dupnet} and adversarial training (AT)~\cite{liu2019extending}. 
For experiments using the three pre-processing defense methods SRS, SOR, and DUP-Net, we generate adversarial examples on a model without defense and then calculate the ASR on a model with defense.
For AT, we adapt the PGD-AT model with the $l_2$ norm constraint and the budget is set as $1$. Please refer to Tab.\ref{tab:defend} for comparative results.
All the detailed settings for the attack methods in Tab.\ref{tab:defend} are the same as in Tab.\ref{tab:main}.
It's worth noting that the adversarial strength of HiT-ADV comes from deformations in the overall shape, which is why we perform well on three preprocessing defense methods. 
We also demonstrate that point-based AT models, while generally effective against point-based attacks, are not effective at defending against shape-based attacks.
This experiment shows that the attack method we proposed further reveals the vulnerability of DNN models. 
The question of how to effectively enhance robustness against shape-based attacks requires further research.

\vspace{-3mm}
\paragraph{Visualization.} The visual comparison of the adversarial examples generated by each method is shown in Fig.~\ref{fig:vis}. 
It is quite evident that, aside from HiT-ADV, other methods more or less have obvious outliers. 
It is an unavoidable common problem of point-based attack methods because it is difficult to generate adversarial examples that are sufficiently adversarial and free of outliers with soft regularization terms in the limited search space constrained by the original shape.
Furthermore, our shape-based perturbation is also sufficiently natural to fully ensure imperceptibility.
Note that the Eps-iso method, which constrains outliers with relatively strong edge length loss and Laplacian loss, claims to be a shape-based attack. 
However, this method firstly does not relax the search space, which still allows outliers to escape constraints, and secondly, the excessive aggregation of point clouds results in the loss of object surface details. 
These problems lead to unsatisfactory visualization results when point clouds are used as input for this method.

\begin{figure*}[t]
  \centering
   \includegraphics[width=1\linewidth]{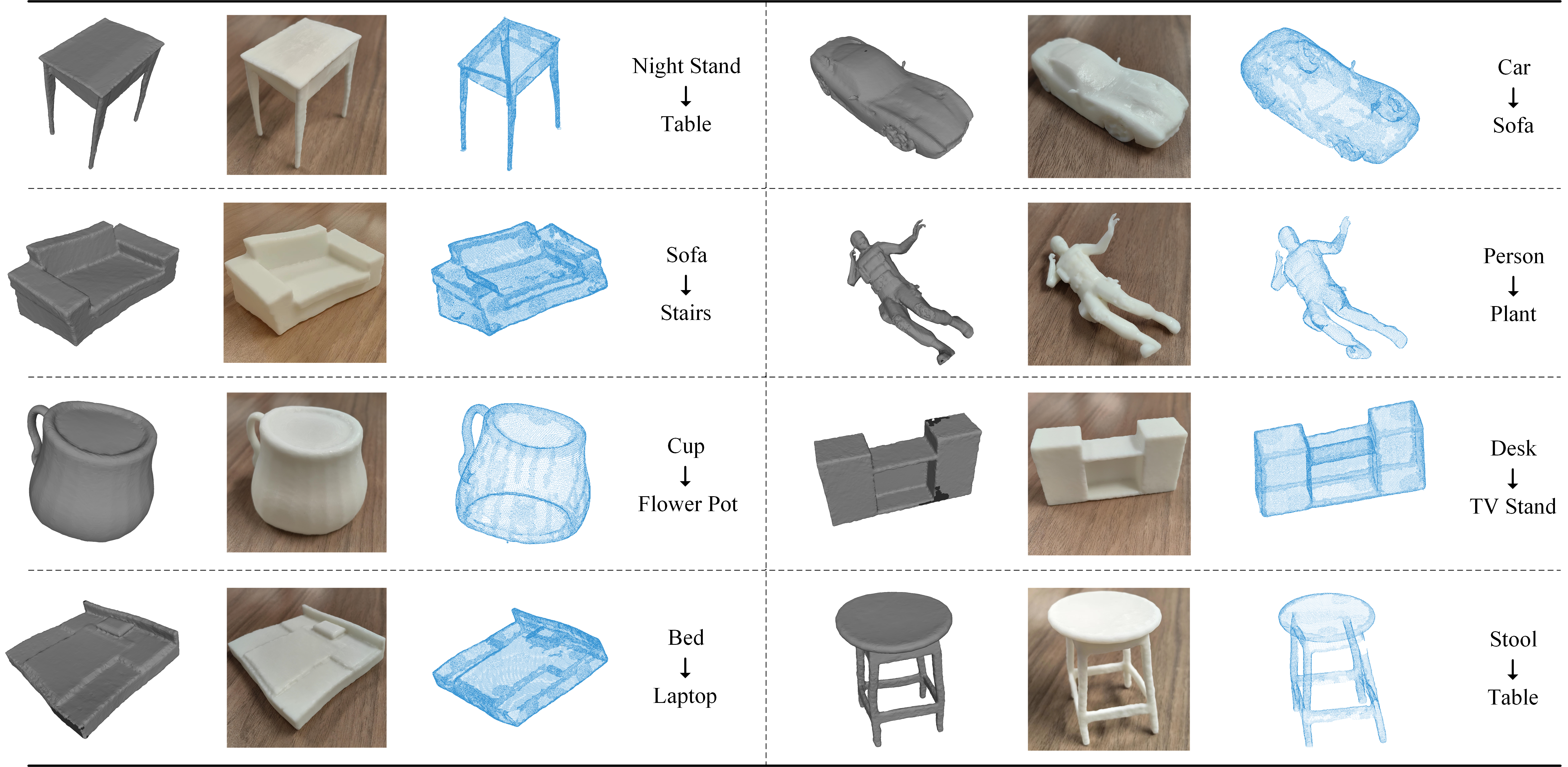}

   \caption{The visualization of the physical adversarial attack process using HiT-ADV, which incorporates benign resampling and benign rigid transformations, is shown in a sequence from left to right: the mesh reconstructed from the adversarial point cloud, the physical adversarial sample produced by 3D printing, the point cloud obtained from rescanning and classification results comparison.}
   \label{fig:physical}
   \vspace{-1mm}
\end{figure*}


\paragraph{Results of Physical Attack.}
In this experiment, we employ the optimization methods discussed in Sec.~\ref{sec:physical} to further enhance the physical adversarial strength.
To validate the effectiveness of adversarial examples generated by HiT-ADV in physical scenarios, we conduct a physical attack experiment and visualize the intermediate results of each stage in Fig.~\ref{fig:physical}.
The overall physical attack process includes reconstructing into a mesh, 3D printing as an adversarial object, and re-scanning to form the input point cloud.
As previously analyzed, in the physical attack process, the digital adversarial strength stemming from outlier points, non-shape-altering point movements, and adversarial rigid transformations can not survive. 
However, HiT-ADV still achieves successful physical attacks, owing to its adversarial strength derived purely from the deformation of object.

\paragraph{Ablation Study.}
We conduct an ablation experiment focusing on the three regularization terms in $\mathcal{L}_{dis}$, with the results shown in Table 1. It is apparent that using all three regularization terms simultaneously can reduce three kinds of distance metrics without sacrificing adversarial strength. 
When any regularization term is absent in $\mathcal{L}_{dis}$, all three distance metrics increase to varying degrees, indicating a decrease in imperceptibility, which proves the necessity of employing all three regularization terms.
Additionally, $\mathcal{L}_{hide}$ notably impacts CSD, indicating its efficacy in hiding deformation perturbations in complex areas. 
For more ablation experiments, please refer to the appendix section.



\begin{table}[t]
\centering

\scalebox{0.8}{
\begin{tabular}{@{}ccc|cccc@{}}
\toprule
$\mathcal{L}_{ker}$       & $\mathcal{L}_{cha}$       & $\mathcal{L}_{hide}$      & ASR $\uparrow$                   & CSD $\downarrow$                 & Uniform $\downarrow$             & KNN $\downarrow$                 \\ \midrule
                          & \checkmark & \checkmark & \textbf{100.00} & 0.6145                           & 0.3542                           & 0.8196                           \\
\checkmark &                           & \checkmark & \textbf{100.00} & 0.5020                           & 0.3412                           & 0.7814                           \\
\checkmark & \checkmark &                           & \textbf{100.00} & 0.8828                           & 0.2945                           & 0.7729                           \\
\checkmark & \checkmark & \checkmark & \textbf{100.00} & \textbf{0.4709} & \textbf{0.2883} & \textbf{0.7447} \\ \bottomrule
\end{tabular}
}
\caption{The results of ablation experiment on different regularization term constraints in $\mathcal{L}_{dis}$. The KNN distances in the table also need to be multiplied by $10^{-3}$.}
\vspace{-3mm}
\label{tab:ablation}
\end{table}

\section{Conclusion}
In this paper, we investigate and analyze the inherent issue of poor imperceptibility in point cloud adversarial attacks.
We devise an imperceptible shape-based point cloud adversarial attack method called HiT-ADV, which conceals deformation perturbations in complex surface areas.
Furthermore, we propose to suppress digital adversarial strength by benign rigid transformations and benign resampling.
Extensive experiments validate the superiority of HiT-ADV, demonstrating its ability to achieve a better trade-off between imperceptibility and adversarial strength and its feasibility to be extended to physical scenarios. Our work demonstrates that the robustness of existing point cloud defense methods is still insufficient, and models remain vulnerable. 
We hope to inspire efforts to enhance true robustness.

{
    \small
    \bibliographystyle{ieeenat_fullname}
    \bibliography{main}
}


\end{document}